\documentclass[runningheads]{llncs}

\usepackage{amsmath,amssymb}
\usepackage{booktabs}
\usepackage{array}
\usepackage{microtype}
\usepackage{graphicx}
\usepackage{tikz}
\usepackage{pgfplots}
\usepackage{url}
\usepackage{hyperref}
\graphicspath{{./figures/}{paper/figures/}}

\usepackage{dblfloatfix}
\usepackage[section]{placeins}

\pgfplotsset{compat=1.18}

\newcommand{\tiledattn}{\textsc{TiledAttention}}
\newcommand{\sdpa}{\textsc{SDPA}}

\begin{document}

\title{\tiledattn: a CUDA Tile \sdpa \space Kernel for PyTorch }
\titlerunning{\tiledattn \space for PyTorch}

\author{Taimur Khan\orcidID{0000-0001-7833-5474}}
\authorrunning{Khan, T.}
\institute{Helmholtz Centre for Environmental Research - UFZ}

\maketitle

\begin{abstract}
\tiledattn{} is a scaled dot-product attention (\sdpa) forward operator for SDPA research on NVIDIA GPUs. Implemented in cuTile Python (TileIR) and exposed as a PyTorch-callable function, it is easier to modify than low-level CUDA templates while retaining realistic behavior via online softmax and tiled $K,V$ streaming. Algorithmically, \tiledattn{} follows the established FlashAttention-style online-softmax formulation; our novelty is the cuTile/TileIR implementation strategy, schedule-level modifiability, and reproducible benchmarking/profiling workflow. The approach is both performant and directly editable at the schedule level from Python (tile shapes, staging, shared-memory layout), enabling rapid, reproducible kernel research without template-heavy CUDA/CUTLASS rewrites. We benchmark \tiledattn{} on an NVIDIA DGX GB10 node with a reproducible harness and compare against PyTorch \sdpa{} (auto-dispatch), explicit unfused baselines (\texttt{torch\_sdpa\_math}, standard eager attention), and forced backend probes (FlashAttention2, EffecientAttention, CuDNN Attention) across sequence length, head dimension, and precision (FP16/BF16). While production fused baselines remain stronger overall, \tiledattn{} delivers large speedups over standard eager attention paths and is available for direct use within PyTorch workflows, providing a practical balance between performance and customizability.
\linebreak
\linebreak
\textbf{Code:} \url{https://github.com/thisistaimur/TiledAttention}
\linebreak
\textbf{Supplementary Material:} \url{https://doi.org/10.5281/zenodo.20119619}

\end{abstract}

\section{Introduction}
Foundation models in both language and vision increasingly rely on long-context attention, making scaled dot-product attention (\sdpa) a recurring bottleneck in training and inference~\cite{devlin2019bert,brown2020language,chowdhery2022palm,touvron2023llama2,dosovitskiy2021vit,radford2021clip}. In the long-sequence regime, attention becomes strongly bandwidth- and locality-sensitive, so kernel schedule quality directly affects throughput.
This focus is also practical for European HPC deployments, where CUDA-capable NVIDIA GPU systems remain common in recent EuroHPC and TOP500 snapshots~\cite{taborsky2025eurohpc,top500highlights2024}.

Tiling partitions the computation into blocks sized for on-chip memory and high data reuse; in attention this corresponds to blockwise score/softmax updates while streaming $K,V$ tiles. Recent CUDA Tile / TileIR and cuTile Python expose these schedule choices at a higher level than template-heavy CUDA development~\cite{nvidia_cuda_tile_overview,nvidia_tile_ir_intro,nvidia_cutile_python_docs}.

For SDPA research, a persistent issue is iteration cost: many high-performance kernels are difficult to modify without deep low-level changes, which slows experimentation with new attention variants. This paper studies \tiledattn{}, an online-softmax tiled \sdpa forward operator expressed as a cuTile Python kernel. We explicitly do not claim a new SDPA algorithm; instead, we contribute an implementation substrate and workflow that make the established algorithm easier to inspect, tune, and reproduce on DGX-class CUDA systems.
The broader literature spans early neural attention, transformer-era SDPA, and efficient alternatives for long context~\cite{bahdanau2014nmtattention,vaswani2017attention,tay2020efficienttransformers,han2021surveytransformers,dao2022flashattention}.

\noindent As sequence length $S$ increases, \sdpa increasingly dominates end-to-end throughput in transformer workloads, which motivates our long-context evaluation focus.

\textbf{Contributions.} We make three contributions:
\begin{itemize}
  \item \textbf{Modifiable cuTile \sdpa kernel:} a forward \sdpa operator expressed as a Python tile program with online softmax updates and no materialization of the full attention matrix.
  \item \textbf{DGX-ready reproducibility workflow:} a measurement suite with explicit warmup, timing, correctness checks, and version pinning, demonstrated on DGX GB10 and portable to other DGX systems.
  \item \textbf{FM-oriented evidence for text and vision regimes:} scaling trends across sequence length $S$, head dimension $D$, and dtype (FP16/BF16), plus sensitivity analysis over key tiling parameters for foundation-model use cases~\cite{devlin2019bert,brown2020language,chowdhery2022palm,touvron2023llama2,dosovitskiy2021vit,radford2021clip}.
\end{itemize}
The rest of the paper is organized as follows: Section~2 summarizes related work and baseline positioning, Section~3 presents the method, Section~4 details implementation and evaluation setup, and Section~5 reports results.

\section{Related Work}
\paragraph{FlashAttention and fused SDPA baselines.}
FlashAttention introduced IO-aware exact attention with online softmax and blockwise streaming~\cite{dao2022flashattention}, and subsequent work improved parallelism and kernel partitioning~\cite{dao2023flashattention2}. In production PyTorch usage, \texttt{torch\_sdpa} auto-dispatch selects among fused backends at runtime; throughout this paper, the forced fused FlashAttention backend we evaluate is referred to as FlashAttention2.

\paragraph{Triton and programmable kernel stacks.}
Triton provides a productive DSL for custom GPU kernels and is a natural comparison point for modifiable attention kernels~\cite{triton}. Relative to Triton, our cuTile/TileIR positioning emphasizes: (i) explicit tile-program control through NVIDIA's tile stack with schedule knobs directly surfaced in our harness, (ii) a PyTorch-facing workflow centered on reproducible benchmark/profiler artifacts for SDPA schedule studies, and (iii) direct alignment with PyTorch auto-dispatch baselines used in production inference/training flows.

\section{Method}
\subsection{Scaled dot-product attention}
Given queries $Q\in\mathbb{R}^{S\times D}$, keys $K\in\mathbb{R}^{S\times D}$, and values $V\in\mathbb{R}^{S\times D}$, \sdpa computes
\begin{equation}
  O = \mathrm{softmax}\left(\frac{QK^\top}{\sqrt{D}} + M\right)V,
  \label{eq:sdpa}
\end{equation}
where $M$ is a mask (e.g., causal or padding). Materializing the score matrix $QK^\top\in\mathbb{R}^{S\times S}$ is prohibitive for long $S$, motivating blockwise implementations.

\subsection{Online softmax and blockwise attention}
The FlashAttention line of work~\cite{dao2022flashattention,dao2023flashattention2} established blockwise, IO-aware attention as a strong baseline. It streams $K,V$ tiles while maintaining running softmax statistics (running max and normalization), avoiding materialization of the $S\times S$ score matrix.

\subsection{Implementations and customization trade-offs}
Production-grade attention kernels are often written in low-level CUDA/CUTLASS-style templates or JIT-compiled DSLs (e.g., Triton). While these achieve high performance, the implementation complexity can make it hard to (i) explore alternative schedules, (ii) isolate bottlenecks, and (iii) reproduce results across environments.

cuTile Python offers a tile-programming model that maps to a CUDA tile IR, keeping the kernel surface area accessible while enabling realistic schedules on modern GPUs. \tiledattn{} uses this model to expose tiling/staging parameters directly in Python, while PyTorch \sdpa{} auto-dispatch remains our production-oriented reference baseline.

\section{Implementation and Evaluation Setup}
We target a systems-oriented kernel study with three explicit goals.
\begin{description}
  \item[G1: Real and modifiable kernel in cuTile.] Implement \sdpa forward with online softmax and masking, not a simplified microkernel, while keeping schedule choices easy to change.
  \item[G2: FM-relevant scaling for text and vision.] Track a strong baseline across shape regimes representative of deployed foundation models, especially $D{=}64$/$128$ and $S\in[512,8192]$~\cite{devlin2019bert,brown2020language,chowdhery2022palm,touvron2023llama2,dosovitskiy2021vit,radford2021clip}.
  \item[G3: Transparent reproducibility.] Provide a benchmark harness and reporting that make results repeatable and interpretable across DGX-class CUDA systems.
\end{description}

We focus on the forward operator because it is (i) a common inference hot spot and (ii) an easier baseline to validate correctness and performance before extending to backward, KV-cache, and fused epilogues.

\subsection{\tiledattn{} Kernel Design}
\subsubsection{Tensor shapes and layout}
We assume inputs $q,k,v$ with shape $[B,H,S,D]$. For kernel execution, we reshape to $[BH,S,D]$ and tile the $S$ dimension. To enable coalesced key loads, we access $K^\top$ (conceptually $K_t\in\mathbb{R}^{BH\times D\times S}$).

\subsubsection{Blockwise online softmax}
Each cooperative thread array (CTA) owns a block of queries $Q_{\mathrm{tile}}\in\mathbb{R}^{T_M\times D}$ and streams tiles of $K,V$ along the sequence dimension. For each $K,V$ tile, the kernel computes partial scores $S_{ij}=\langle q_i,k_j\rangle/\sqrt{D}$, applies a mask, and updates the online softmax state.

We maintain, per query row $i$ in the CTA tile:
\begin{align}
  m_i &\leftarrow \max(m_i, \max_j S_{ij}),\\
  \ell_i &\leftarrow \ell_i\cdot e^{m_i^{\mathrm{old}}-m_i} + \sum_j e^{S_{ij}-m_i},\\
  o_i &\leftarrow o_i\cdot e^{m_i^{\mathrm{old}}-m_i} + \sum_j e^{S_{ij}-m_i} v_j,
\end{align}
where $m_i$ is the running max, $\ell_i$ is the running normalizer, and $o_i\in\mathbb{R}^{D}$ is the running output accumulator. After streaming all tiles, we write $O_i=o_i/\ell_i$.

\subsubsection{Numerical choices}
Inputs are FP16 or BF16. We accumulate dot products, softmax statistics, and output accumulation in FP32 for stability (matching common practice in high-performance attention kernels). We optionally support causal and padding masks; causal masking avoids reading or contributing to tiles strictly above the diagonal.

\begin{figure}[!htb]
  \centering
  \small
  \begin{tabular}{c c c c c}
    \fbox{Load $Q$ tile} &
    $\rightarrow$ &
    \fbox{Stream $K,V$ tiles + score/mask} &
    $\rightarrow$ &
    \fbox{Update $(m,\ell,o)$; normalize + store}
  \end{tabular}
  \caption{Figure 2: \tiledattn{} forward pipeline at a glance.}
  \label{fig:pipeline}
\end{figure}

\begin{table}[!htb]
\centering
\caption{Tuning/search parameters exposed by \tiledattn{} (representative).}
\label{tab:tuning}
\small
\setlength{\tabcolsep}{4pt}
\renewcommand{\arraystretch}{0.95}
\begin{tabular}{@{}p{2.3cm}p{3.2cm}p{5.3cm}@{}}
\toprule
Parameter & Typical values & Effect \\
\midrule
$T_M$ (Q rows) & 64, 128 & CTA work per launch; affects occupancy and reuse \\
$T_N$ (KV columns) & 64, 128, 256 & Streaming granularity; impacts memory coalescing and softmax update cost \\
Stages / prefetch & 1--4 & Hides global memory latency via pipelining \\
Shared-mem layout & row/col swizzles & Impacts bank conflicts and vectorized loads \\
\bottomrule
\end{tabular}
\end{table}

\subsection{Implementation}
\subsubsection{API}
We expose a minimal functional API:
\begin{equation}
  \texttt{sdpa(q, k, v, causal, scale)} \rightarrow \texttt{o}.
\end{equation}
The API matches common frameworks: $\texttt{q,k,v}$ are contiguous tensors in $[B,H,S,D]$, \texttt{causal} selects causal masking, and \texttt{scale} defaults to $1/\sqrt{D}$.

\paragraph{PyTorch entry point.}
\tiledattn{} is available as a Python package that integrates with PyTorch. A typical usage pattern is:
\begin{verbatim}
import torch
from tiledattention import sdpa

# q,k,v must be CUDA tensors with dtype float16 or bfloat16.
o = sdpa(q, k, v, causal=True)  # q,k,v: [B, H, S, D]
\end{verbatim}

\subsubsection{Compilation and caching}
The cuTile kernel is JIT-compiled and cached by a key $(T_M,T_N,D,\mathrm{dtype},\mathrm{causal})$, avoiding recompilation and ensuring steady-state measurement.

\subsubsection{Benchmark harness}
We evaluate kernels using a reproducible harness with the following policies:
\begin{itemize}
  \item \textbf{Warmup:} run $N_w$ iterations to trigger compilation and stabilize clocks.
  \item \textbf{Timing:} CUDA events around the forward call; report median and p95 over $N_r$ repetitions.
  \item \textbf{Correctness:} for small shapes (e.g., $S\le 256$), compare to a high-precision reference (FP32) within a tolerance that accounts for FP16/BF16.
  \item \textbf{Isolation:} synchronize before timing, avoid interleaving other GPU work.
\end{itemize}
All runs are executed on a DGX GB10 node; the same scripts apply to other DGX systems with CUDA-compatible stacks.

\begin{table}[!htb]
\centering
\caption{Reproducibility checklist for the measured runs in this artifact.}
\label{tab:repro}
\small
\setlength{\tabcolsep}{4pt}
\renewcommand{\arraystretch}{0.95}
\begin{tabular}{@{}p{2.7cm}p{8.0cm}@{}}
\toprule
Item & Value \\
\midrule
System & NVIDIA DGX GB10 host GPU \\
GPU driver & 580.126.09 \\
CUDA toolkit & CUDA 13.1 (V13.1.115) \\
cuTile & 1.1.0 \\
PyTorch & 2.10.0+cu130 \\
Timing method & CUDA events; median + p95 over $N_r$ runs \\
Profiler & NVIDIA Nsight \\ 
\bottomrule
\end{tabular}
\end{table}

\subsection{Experimental Setup}
We sweep a workload grid intended to cover common foundation-model attention shapes:
\begin{itemize}
  \item Sequence length $S\in\{512, 1024, 2048, 4096, 8192\}$.
  \item Head dimensions $D\in\{64, 96, 128, 160\}$.
  \item Dtypes: FP16 and BF16.
  \item Masking: causal and non-causal.
\end{itemize}
This grid is anchored in common transformer deployments: $D{=}64$ is common in BERT- and ViT-style models~\cite{devlin2019bert,dosovitskiy2021vit,radford2021clip}, while $D{=}128$ is prevalent in larger decoder-only language models such as GPT-3, PaLM, and LLaMA-family systems~\cite{brown2020language,chowdhery2022palm,touvron2023llama2}. Sequence lengths 512--2048 cover short-to-mid context workloads common in serving and fine-tuning, while 4096--8192 capture long-context pressure in modern FM deployments.

We compare \tiledattn{} to PyTorch \sdpa{} on the same node with identical tensor initialization, streams, and timing methodology. We use \texttt{torch\_sdpa} (PyTorch auto-dispatch) as the primary baseline because it selects the most suitable fused backend available at runtime for each shape/dtype. For deeper analysis, we additionally probe explicit unfused baselines (\texttt{torch\_sdpa\_math} and standard eager attention) and forced fused backend variants (FlashAttention2, EffecientAttention, CuDNN Attention) in the benchmark artifacts. A fully tuned Triton SDPA baseline is outside the scope of this study because it would require separate kernel engineering and shape-specific tuning across the full grid; without equal tuning effort across frameworks, the comparison would be difficult to interpret fairly.

\section{Results}
\begin{figure}[!htb]
\centering
\includegraphics[width=0.95\linewidth]{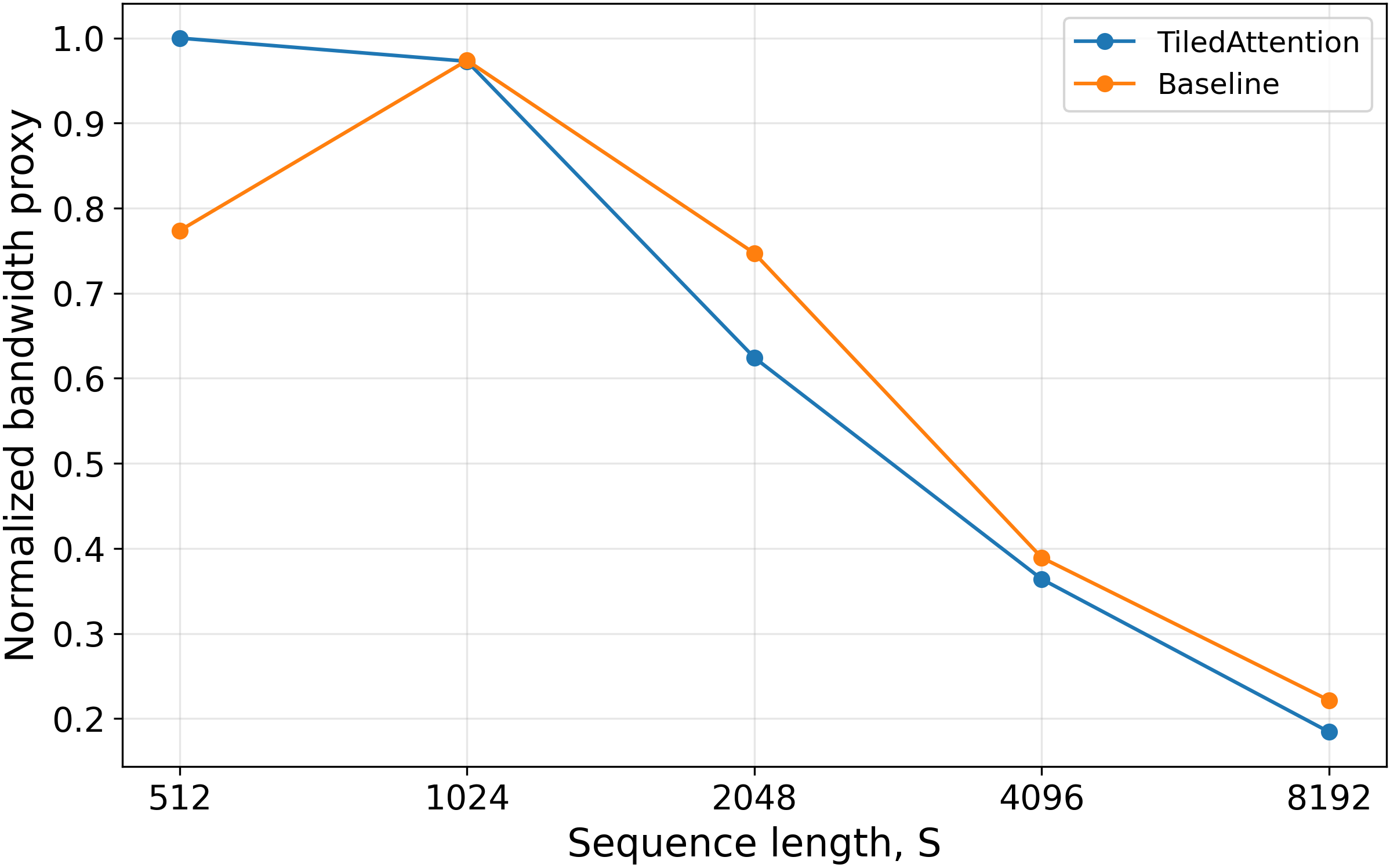}
\caption{Normalized bandwidth proxy versus sequence length (FP16, $D{=}128$, non-causal).}
\label{fig:bottleneck}
\end{figure}

The full results of the benchmark with summaries and NSight~\cite{nvidiaNsightDeep} logs are available in the Supplementary Material~\cite{tiledattention-supplementary}.
We report (i) time per forward $t_{\mathrm{fwd}}$ (ms), (ii) throughput normalized to tokens/s, and (iii) a normalized bandwidth proxy for Figure~\ref{fig:bottleneck}.
\begin{equation}
  \mathrm{Throughput} = \frac{B\cdot H\cdot S}{t_{\mathrm{fwd}}} \quad \text{(tokens/s)}.
\end{equation}
For the bandwidth proxy we first compute
\begin{equation}
  \mathrm{BW}_{\mathrm{proxy}} = \frac{4\cdot B\cdot H\cdot S\cdot D\cdot \mathrm{bytes\_per\_elem}}{t_{\mathrm{fwd}}},
\end{equation}
then normalize by the maximum value among plotted methods within the same figure panel.
Here, $B$ is batch size, $H$ is number of heads, $S$ is sequence length, $D$ is per-head channel dimension, \(\mathrm{bytes\_per\_elem}\) is bytes per tensor element for the active dtype, $t_{\mathrm{fwd}}$ is measured forward-pass time, and $N_r$/$N_w$ denote timed/warmup iteration counts.
For forced backend probes that are unavailable for a given shape/dtype, we record \texttt{NaN} metrics and annotate status in the CSV outputs (\texttt{status}, \texttt{status\_detail}) instead of dropping those rows.
Unless stated otherwise, point estimates in plots are medians over $N_r$ timed iterations after $N_w$ warmup; p95 values and per-run traces are available in supplementary CSV artifacts.

\subsection{Throughput scaling vs sequence length}
\begin{figure}[!htb]
\centering
\includegraphics[width=0.85\linewidth]{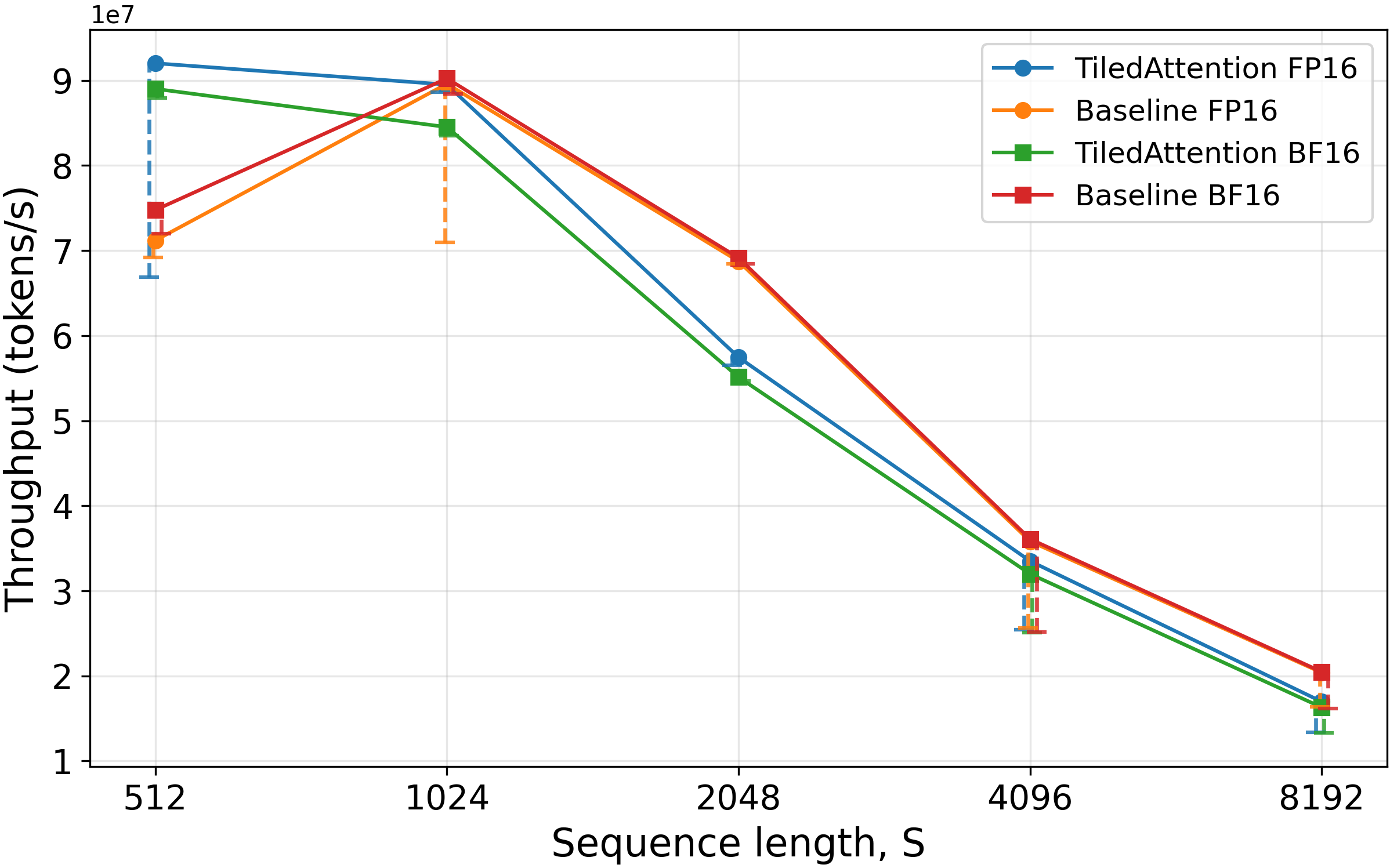}
\caption{Throughput versus sequence length for $D{=}128$ (FP16/BF16, non-causal). Points show median throughput; downward error bars show the p95-latency-equivalent throughput bound (whiskers are slightly x-offset for readability).}
\label{fig:throughput}
\end{figure}

Figure~\ref{fig:throughput} plots throughput versus $S$ for FP16 and BF16; long-$S$ trends reflect memory traffic, while short-$S$ is overhead- and utilization-limited.
In this paper, we refer to $S\in\{512,1024,2048\}$ as short-to-mid context and $S\in\{4096,8192\}$ as long context.
Figure~\ref{fig:throughput} shows median point estimates with one-sided p95-derived error bars (downward whiskers in throughput space); other main plots use median point estimates only, with p95 and per-run variability provided in the supplementary CSV artifacts.

\paragraph{Observed behavior in our runs.}
Across the full study grid (80 points: $S\in\{512,1024,2048,4096,8192\}$, $D\in\{64,96,128,160\}$, FP16/BF16, causal/non-causal), \tiledattn{} achieves a mean throughput ratio of $0.632\times$ versus PyTorch \sdpa{} (auto-dispatch) (median $0.634\times$), with 4 wins out of 80 points. The closest regime is $D{=}128$ (mean $0.947\times$, 4/20 wins). For $D\in\{64,96,160\}$, auto-dispatch \sdpa{} remains faster on average.
We also observe a transition around $S\approx 2048$: short-$S$ points can approach parity, while long-$S$ becomes more sensitive to memory traffic and instruction mix. For example, in FP16 non-causal at $D{=}128$, \tiledattn{} reaches $73.62$ TFLOP/s at $S{=}4096$ versus $78.98$ for auto-dispatch \sdpa{}, and $73.38$ versus $88.84$ at $S{=}8192$.

\subsection{Explicit baseline summary}
\begin{figure}[!htb]
\centering
\includegraphics[width=1\linewidth]{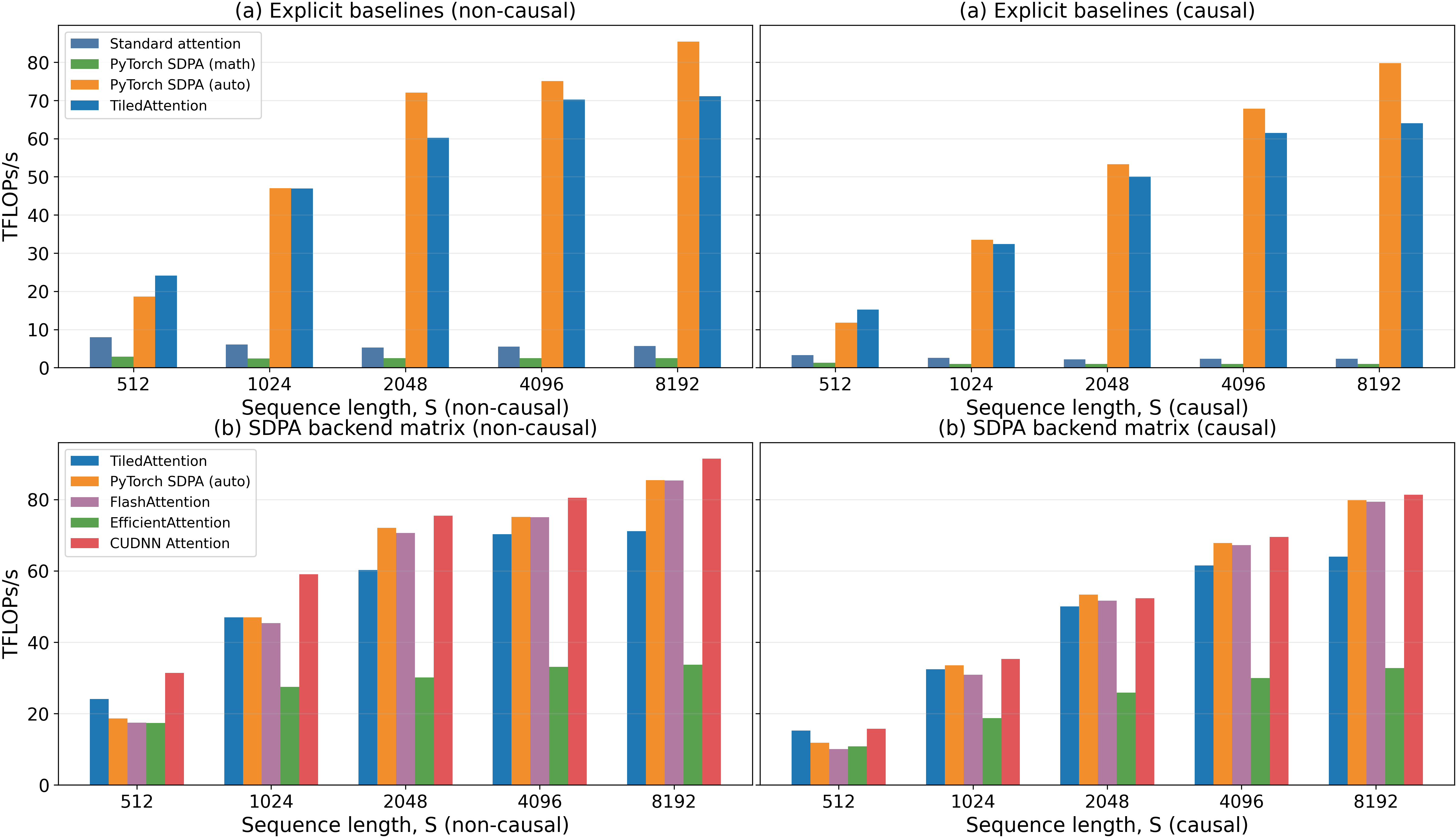}
\caption{Composite explicit-baseline view (FP16, $D{=}128$). Top row (a): \tiledattn{} vs \texttt{torch\_sdpa} (auto), \texttt{torch\_sdpa\_math}, and standard eager attention. Bottom row (b): \tiledattn{} vs PyTorch \sdpa{} backend matrix (\texttt{torch\_sdpa} auto, FlashAttention2, EffecientAttention, CuDNN Attention). Unsupported backend points are recorded as NaN in the benchmark CSV.}
\label{fig:explicit-baselines}
\end{figure}

To make the value proposition explicit for HPC users, we report \tiledattn{} not only against PyTorch \sdpa{} (auto-dispatch), but also against unfused baselines. Table~\ref{tab:explicit-baselines} shows that while auto-dispatch \sdpa{} is still the strongest overall baseline, \tiledattn{} provides large speedups over standard eager attention and PyTorch's math \sdpa{} path across the full grid. Figure~\ref{fig:explicit-baselines}(a) visualizes this explicit-baseline comparison, while Figure~\ref{fig:explicit-baselines}(b) provides a backend-level view of PyTorch \sdpa{} at $D{=}128$.
In Figure~\ref{fig:explicit-baselines}(b), PyTorch auto-dispatch generally tracks the strongest available fused backend for each point; FlashAttention2 is typically closest to auto-dispatch on these runs, EffecientAttention is generally lower on this workload, and CuDNN Attention is competitive for some settings but more shape-sensitive.

\begin{table}[!htb]
\centering
\caption{Aggregate throughput summary over 80 study points (tokens/s ratio).}
\label{tab:explicit-baselines}
\small
\setlength{\tabcolsep}{3.5pt}
\renewcommand{\arraystretch}{0.95}
\begin{tabular}{@{}p{3.4cm}p{2.0cm}p{2.0cm}p{1.8cm}@{}}
\toprule
Comparison & Mean ratio & Median ratio & Win count \\
\midrule
\tiledattn{} / PyTorch \sdpa{} (auto) & 0.632$\times$ & 0.634$\times$ & 4 / 80 \\
\tiledattn{} / PyTorch \sdpa{} (math) & 28.15$\times$ & 20.31$\times$ & 80 / 80 \\
\tiledattn{} / standard eager attention & 14.36$\times$ & 10.76$\times$ & 80 / 80 \\
\bottomrule
\end{tabular}
\end{table}

\subsection{Regime map over $(S, D)$}
\begin{figure}[!htb]
\centering
\includegraphics[width=0.85\linewidth]{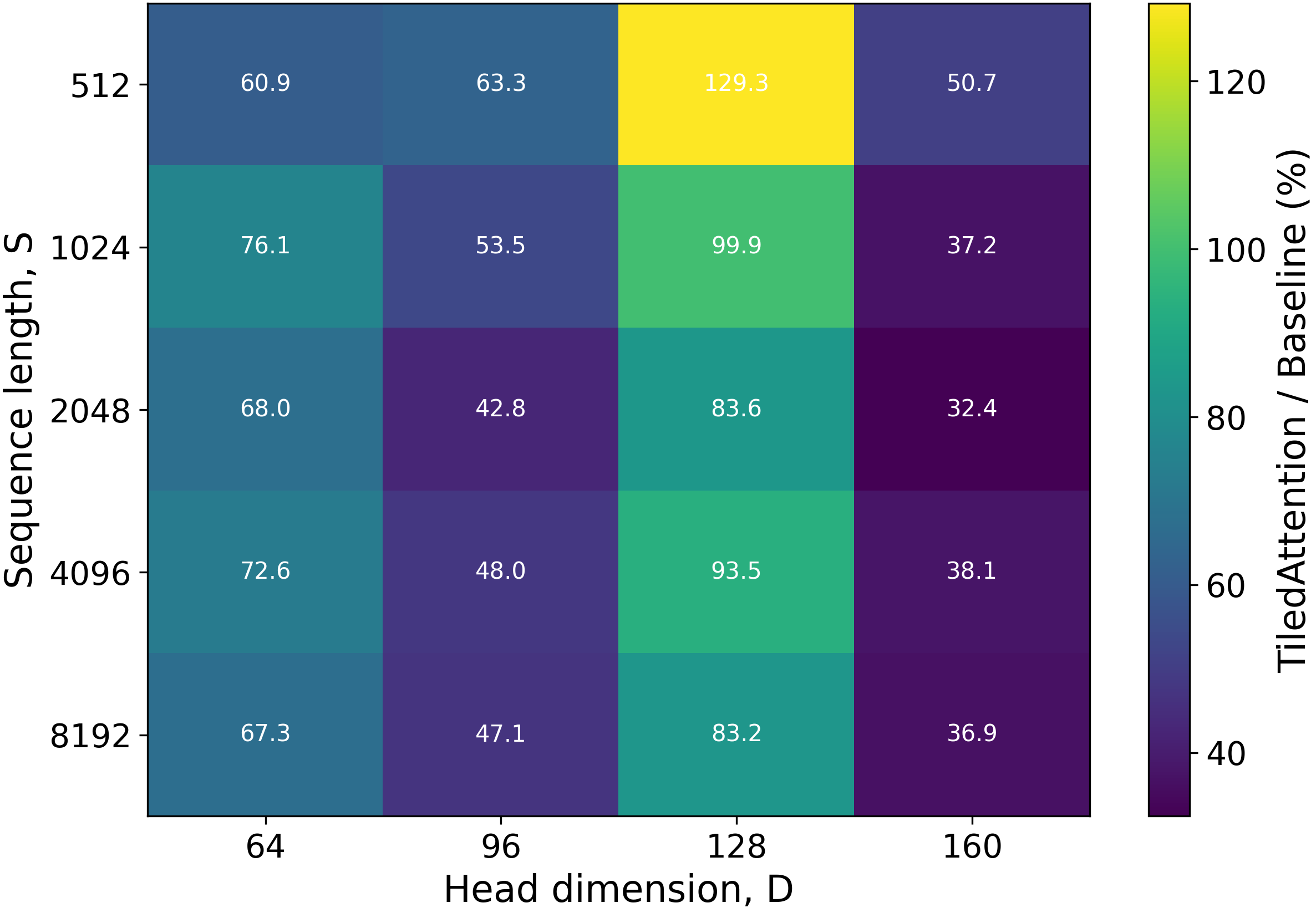}
\caption{Relative performance regime map (\tiledattn{} as \% of PyTorch \sdpa{} (auto-dispatch)) for FP16 non-causal runs.}
\label{fig:heatmap}
\end{figure}

A regime heatmap is a compact way to summarize which shapes are competitive. Figure~\ref{fig:heatmap} illustrates the intended presentation: cells show \tiledattn{} performance as a percentage of PyTorch \sdpa{} (auto-dispatch) (100\% is parity).

In the measured regime map, the largest gaps appear at higher or non-power-of-two head dimensions (notably $D{=}160$ and $D{=}96$), while $D{=}128$ remains closest to parity. Averaged over all sequence lengths, dtypes, and masking modes, throughput ratios (\tiledattn{}/PyTorch \sdpa{} auto-dispatch) are: $D{=}64:0.727\times$, $D{=}96:0.513\times$, $D{=}128:0.947\times$, $D{=}160:0.343\times$.
The largest gap appears at $D{=}160$ (mean ratio $0.343\times$), where register/shared-memory pressure and reduced warp-level utilization are most pronounced in our current schedule. Two practical mitigation paths are: (i) dimension-aware tile/staging policies (e.g., smaller $T_N$ and/or adjusted staging at high $D$), and (ii) targeted kernel variants for non-power-of-two or larger head dimensions to reduce bank conflicts and register pressure.

\subsection{Profiling-guided bottleneck analysis}
To interpret scaling trends, we profile representative shapes with Nsight Compute (counters) and Nsight Systems (timeline/API overhead). We track throughput, memory traffic, and stalls to identify when the kernel becomes memory-bound, what limits short-$S$, and which knobs move these regimes.

In our experience, long-$S$ shapes show high shared- and global-memory pressure, so staging depth and shared-memory layout can dominate. Short-$S$ shapes can be limited by insufficient parallelism and overheads, where larger $T_M$ may help at the cost of occupancy.

For non-causal FP16 at $B{=}1,H{=}8,S{=}4096,D{=}128$, one-pass Nsight Compute shows primary-kernel times of $1.203$ ms (\tiledattn{}), $1.132$ ms (PyTorch \sdpa{} auto), and $1.131$ ms (forced FlashAttention2). For causal $S{=}4096,D{=}128$, the corresponding times are $0.6855$ ms, $0.6124$ ms, and $0.5912$ ms. This is consistent with the backend-matrix behavior in Figure~\ref{fig:explicit-baselines}(b): auto-dispatch is very close to forced FlashAttention2, while \tiledattn{} remains competitive but slower on this shape.
Across these Nsight runs, achieved warp activity remains below 30\% of peak for all three methods (roughly 8--13\%), with \tiledattn{} showing higher DRAM/L2 throughput percentages than auto/forced FlashAttention on the profiled shape. This indicates that long-$S$ performance is primarily constrained by memory movement and pipeline utilization, with additional headroom from schedule-level tuning.
For reproducibility, we provide raw profiler artifacts and summaries in the supplementary materials: \texttt{\path{ncu_*.ncu-rep}}, \texttt{\path{ncu_*_raw.csv}}, and \texttt{\path{ncu_profile_summary_*.md}}; timeline traces are included as \texttt{\path{*.nsys-rep}} when collected.

\subsection{Sensitivity to tiling parameters}
A central advantage of expressing the kernel as a tile program is the ability to expose and sweep tiling parameters. Table~\ref{tab:besttiles} summarizes the best-performing tile settings by regime and reports sensitivity: how far performance drops when using a non-optimal but reasonable tile choice.

\begin{table}[!htb]
\centering
\caption{Best tile settings by regime from the measured tuning sweep.}
\label{tab:besttiles}
\small
\setlength{\tabcolsep}{3.5pt}
\renewcommand{\arraystretch}{0.95}
\begin{tabular}{@{}p{2.0cm}p{2.3cm}p{2.0cm}p{2.0cm}p{1.5cm}@{}}
\toprule
Regime & Shapes & Best $(T_M,T_N)$ & Runner-up & Sensitivity \\
\midrule
Short $S$ & $S{=}1024, D{=}128$ & (64, 128) & (64, 64) & 3.92\% \\
Mid $S$ & $S{=}4096, D{=}128$ & (64, 128) & (64, 64) & 8.53\% \\
Long $S$ & $S{=}8192, D{=}128$ & (64, 128) & (64, 64) & 9.39\% \\
\bottomrule
\end{tabular}
\end{table}

\subsection{Optimization ladder on reduced benchmark}
To show what the tile-programming workflow provides in practice, Table~\ref{tab:opt-ladder} reports a focused reduced benchmark before/after schedule-policy refinements. The key point is not a universal win from one static configuration, but that a shape-aware policy recovers performance across mixed regimes without low-level CUDA rewrites.

\begin{table}[!htb]
\centering
\caption{Reduced benchmark (\{(1024,64), (2048,64), (4096,128)\}, FP16, non-causal): ratio to PyTorch \sdpa{} (auto-dispatch) throughput.}
\label{tab:opt-ladder}
\small
\setlength{\tabcolsep}{3.5pt}
\renewcommand{\arraystretch}{0.95}
\begin{tabular}{@{}p{3.4cm}p{1.5cm}p{1.5cm}p{1.5cm}@{}}
\toprule
Config & $(1024,64)$ & $(2048,64)$ & $(4096,128)$ \\
\midrule
\texttt{async\_default} (fp32) & 0.766$\times$ & 0.668$\times$ & 1.017$\times$ \\
\texttt{tm64\_tn64} (fp16acc) & 0.779$\times$ & 0.698$\times$ & 0.968$\times$ \\
\texttt{async\_auto} (shape-aware) & 0.775$\times$ & 0.694$\times$ & 1.017$\times$ \\
\bottomrule
\end{tabular}
\end{table}

\section{Discussion}
Across our runs, two knobs dominate first-order behavior: tile sizes $(T_M,T_N)$ and staging depth. Larger tiles can improve reuse and streaming efficiency but may reduce occupancy through shared-memory and register pressure. Reliable comparison therefore requires a stable harness (warmup, isolated stream, median/p95 reporting), especially with JIT-compiled kernels.

\paragraph{When should HPC users adopt \tiledattn{}?}
Table~\ref{tab:when-to-use} summarizes a pragmatic deployment view. The primary advantage is controllability and iteration speed for architecture-specific tuning and research, while PyTorch \sdpa{} auto-dispatch remains the default for maximum out-of-the-box throughput.

\paragraph{Strong competitors make custom-kernel value clearer.}
FlashAttention2-style fused kernels~\cite{dao2022flashattention,dao2023flashattention2} remain a strong external competitor and set a high bar for attention performance. Triton-based kernels are also compelling for custom-kernel development~\cite{triton}. Against that context, the value of \tiledattn{} is practical: fast kernel iteration, clear schedule controls, and reproducible profiling for architecture-specific tuning in HPC settings.

\begin{table}[!htb]
\centering
\caption{Decision guide for HPC deployment.}
\label{tab:when-to-use}
\small
\setlength{\tabcolsep}{4pt}
\renewcommand{\arraystretch}{0.95}
\begin{tabular}{@{}p{2.7cm}>{\centering\arraybackslash}p{3.8cm}p{3.6cm}@{}}
\toprule
Priority & Recommended path & Rationale \\
\midrule
Peak production throughput across broad shapes & PyTorch \sdpa{} (auto) & Best aggregate performance in this study. \\
Rapid kernel iteration / schedule research & \tiledattn{} & Python-level tile policy changes with reproducible profiling loop. \\
Custom kernel development in a broader DSL ecosystem & Triton / FlashAttention codebases & Strong alternatives when teams can afford separate kernel engineering and tuning effort. \\
Custom masking/layout experiments & \tiledattn{} & Easier to modify kernel behavior than template-heavy CUDA paths. \\
Replacing unfused attention baselines & \tiledattn{} & Large gains vs eager and math-\sdpa{} paths (Table~\ref{tab:explicit-baselines}). \\
\bottomrule
\end{tabular}
\end{table}

\paragraph{Why short-$S$ can win while long-$S$ loses.}
At short sequence lengths, tuned tile shapes can deliver good locality and low overhead, yielding isolated wins. As $S$ increases, streamed-tile work and online-softmax update cost grow, and occupancy/instruction-mix limitations become more visible; fused production \sdpa{} kernels therefore retain higher sustained long-context throughput.

\section{Limitations and Future Work}
This work has three main limitations. First, we evaluate only the forward pass; backward support and KV-cache-oriented kernels are future work needed for full training/serving parity. Second, we target Grace--Blackwell / Blackwell-class GPUs; portability to other NVIDIA generations and non-NVIDIA vendors may require additional schedules and backend-specific adaptations. Third, we study a deliberately small tuning space; extending to larger search spaces and automated policy selection is an important next step.
Concretely, our near-term roadmap is: (i) backward-pass kernels with matching reproducibility harness support, (ii) KV-cache and decoding-oriented variants, and (iii) cross-architecture validation on at least one additional GPU generation.

\section{Conclusion}
We presented \tiledattn{}, a tiled online-softmax \sdpa forward operator expressed as a cuTile Python tile program on Grace--Blackwell. While the algorithmic core follows established FlashAttention-style online softmax, our contribution is an implementation and evaluation workflow that makes schedule-level SDPA experimentation practical and reproducible in Python. Beyond a kernel implementation, we contributed a reproducible benchmark harness and an analysis that highlights how performance scales with sequence length, head dimension, dtype, and tiling choices. Our results support a practical workflow for performance engineering of foundation-model primitives on HPC systems.

\bibliographystyle{splncs04}
\bibliography{references}

@article{vaswani2017attention,
  author  = {Ashish Vaswani and Noam Shazeer and Niki Parmar and Jakob Uszkoreit and Llion Jones and Aidan N. Gomez and Lukasz Kaiser and Illia Polosukhin},
  title   = {Attention Is All You Need},
  journal = {Advances in Neural Information Processing Systems},
  year    = {2017}
}

@article{tay2020efficienttransformers,
  title   = {Efficient Transformers: A Survey},
  author  = {Tay, Yi and Dehghani, Mostafa and Bahri, Dara and Metzler, Donald},
  journal = {arXiv preprint arXiv:2009.06732},
  year    = {2020},
  url     = {https://arxiv.org/abs/2009.06732}
}

@article{han2021surveytransformers,
  title   = {A Survey of Transformers},
  author  = {Han, Kai and Xiao, A and Wu, Enhua and Guo, Jianyuan and Xu, Chao and Wang, Yunhe},
  journal = {arXiv preprint arXiv:2106.04554},
  year    = {2021},
  url     = {https://arxiv.org/abs/2106.04554}
}

@article{bahdanau2014nmtattention,
  title   = {Neural Machine Translation by Jointly Learning to Align and Translate},
  author  = {Bahdanau, Dzmitry and Cho, Kyunghyun and Bengio, Yoshua},
  journal = {arXiv preprint arXiv:1409.0473},
  year    = {2014},
  url     = {https://arxiv.org/abs/1409.0473}
}

@inproceedings{devlin2019bert,
  title     = {BERT: Pre-training of Deep Bidirectional Transformers for Language Understanding},
  author    = {Devlin, Jacob and Chang, Ming-Wei and Lee, Kenton and Toutanova, Kristina},
  booktitle = {Proceedings of NAACL-HLT},
  year      = {2019}
}

@article{brown2020language,
  title   = {Language Models are Few-Shot Learners},
  author  = {Brown, Tom B. and Mann, Benjamin and Ryder, Nick and Subbiah, Melanie and Kaplan, Jared and Dhariwal, Prafulla and Neelakantan, Arvind and Shyam, Pranav and Sastry, Girish and Askell, Amanda and others},
  journal = {Advances in Neural Information Processing Systems},
  year    = {2020}
}

@article{chowdhery2022palm,
  title   = {PaLM: Scaling Language Modeling with Pathways},
  author  = {Chowdhery, Aakanksha and Narang, Sharan and Devlin, Jacob and Bosma, Maarten and Mishra, Gaurav and Roberts, Adam and Barham, Paul and Chung, Hyung Won and Sutton, Charles and Gehrmann, Sebastian and others},
  journal = {arXiv preprint arXiv:2204.02311},
  year    = {2022},
  url     = {https://arxiv.org/abs/2204.02311}
}

@article{touvron2023llama2,
  title   = {Llama 2: Open Foundation and Fine-Tuned Chat Models},
  author  = {Touvron, Hugo and Martin, Louis and Stone, Kevin and Albert, Peter and Almahairi, Amjad and Babaei, Yasmine and Bashlykov, Nikolay and Batra, Soumya and Bhargava, Prajjwal and Bhosale, Shruti and others},
  journal = {arXiv preprint arXiv:2307.09288},
  year    = {2023},
  url     = {https://arxiv.org/abs/2307.09288}
}

@inproceedings{dosovitskiy2021vit,
  title     = {An Image is Worth 16x16 Words: Transformers for Image Recognition at Scale},
  author    = {Dosovitskiy, Alexey and Beyer, Lucas and Kolesnikov, Alexander and Weissenborn, Dirk and Zhai, Xiaohua and Unterthiner, Thomas and Dehghani, Mostafa and Minderer, Matthias and Heigold, Georg and Gelly, Sylvain and Uszkoreit, Jakob and Houlsby, Neil},
  booktitle = {International Conference on Learning Representations (ICLR)},
  year      = {2021},
  url       = {https://arxiv.org/abs/2010.11929}
}

@article{radford2021clip,
  title   = {Learning Transferable Visual Models From Natural Language Supervision},
  author  = {Radford, Alec and Kim, Jong Wook and Hallacy, Chris and Ramesh, Aditya and Goh, Gabriel and Agarwal, Sandhini and Sastry, Girish and Askell, Amanda and Mishkin, Pamela and Clark, Jack and Krueger, Gretchen and Sutskever, Ilya},
  journal = {arXiv preprint arXiv:2103.00020},
  year    = {2021},
  url     = {https://arxiv.org/abs/2103.00020}
}

@misc{triton,
  author = {{OpenAI}},
  title  = {Triton: an intermediate language and compiler for writing efficient GPU kernels},
  year   = {2024},
  note   = {Accessed 2026-01-31}
}

@article{dao2022flashattention,
  title   = {FlashAttention: Fast and Memory-Efficient Exact Attention with IO-Awareness},
  author  = {Dao, Tri and Fu, Daniel Y. and Ermon, Stefano and Rudra, Atri and R{\'e}, Christopher},
  journal = {arXiv preprint arXiv:2205.14135},
  year    = {2022},
  url     = {https://arxiv.org/abs/2205.14135}
}

@article{dao2023flashattention2,
  title   = {FlashAttention-2: Faster Attention with Better Parallelism and Work Partitioning},
  author  = {Dao, Tri},
  journal = {arXiv preprint arXiv:2307.08691},
  year    = {2023},
  url     = {https://arxiv.org/abs/2307.08691}
}

@misc{nvidia_cutile_python_docs,
  title        = {cuTile Python Documentation},
  howpublished = {NVIDIA Documentation},
  url          = {https://docs.nvidia.com/cuda/cutile-python/},
  note         = {Accessed 2026-01-31}
}

@misc{nvidia_tile_ir_intro,
  title        = {Tile IR — Introduction},
  howpublished = {NVIDIA Documentation},
  url          = {https://docs.nvidia.com/cuda/tile-ir/latest/sections/introduction.html},
  note         = {Accessed 2026-01-31}
}

@misc{nvidia_cuda_tile_overview,
  title        = {NVIDIA CUDA Tile},
  howpublished = {NVIDIA Developer},
  url          = {https://developer.nvidia.com/cuda/tile},
  note         = {Accessed 2026-01-31}
}

@article{taborsky2025eurohpc,
  title   = {Towards a European HPC/AI ecosystem},
  journal = {Procedia Computer Science},
  year    = {2025},
  doi     = {10.1016/j.procs.2025.02.269},
  url     = {https://doi.org/10.1016/j.procs.2025.02.269}
}

@misc{top500highlights2024,
  author = {{TOP500}},
  title  = {TOP500 June 2024 Highlights},
  year   = {2024},
  url    = {https://www.top500.org/lists/top500/2024/06/highs/},
  note   = {Accessed 2026-02-16}
}

@misc{nvidiaNsightDeep,
	author = {},
	title = {{N}sight {D}eep {L}earning {D}esigner --- developer.nvidia.com},
	howpublished = {\url{https://developer.nvidia.com/nsight-dl-designer}},
	year = {},
	note = {[Accessed 26-02-2026]},
}

@misc{tiledattention-supplementary,
  doi = {10.5281/zenodo.20119619},
  url = {https://doi.org/10.5281/zenodo.20119619},
  author = {Khan,  Taimur},
  title = {TiledAttention on NVIDIA DGX GB10: Supplementary Benchmark and Nsight Compute Results},
  publisher = {Zenodo},
  year = {2026},
  copyright = {Creative Commons Attribution 4.0 International}
}

\end{document}